\documentclass[a4paper, 10pt, conference]{ieeeconf}

\IEEEoverridecommandlockouts
\usepackage{graphicx}
\usepackage{subcaption}
\usepackage{lipsum}
\usepackage[utf8]{inputenc}

\usepackage{amsmath, amsthm, amsfonts}

\title{\LARGE \bf
Mobile Robots through Task-Based Human Instructions using Incremental Curriculum Learning
}

\author{Muhammad A. Muttaqien$^{1}$, Ayanori Yorozu$^{2}$ and Akihisa Ohya$^{3}$%
\thanks{$^{1}$ Faculty of Systems and Information Engineering, Computer Science,
        University of Tsukuba, Tennodai, Tsukuba, Japan
        {\tt\small muttaqien-m@roboken.cs.tsukuba.ac.jp}}%
\thanks{$^{2, 3}$ Faculty of Systems and Information Engineering, Computer Science,
        University of Tsukuba, Tennodai, Tsukuba, Japan}%
}

\begin{document}

\maketitle
\thispagestyle{empty}
\pagestyle{empty}

\begin{abstract}

This paper explores the integration of incremental curriculum learning (ICL) with deep reinforcement learning (DRL) techniques to facilitate mobile robot navigation through task-based human instruction. By adopting a curriculum that mirrors the progressive complexity encountered in human learning, our approach systematically enhances robots' ability to interpret and execute complex instructions over time. We explore the principles of DRL and its synergy with ICL, demonstrating how this combination not only improves training efficiency but also equips mobile robots with the generalization capability required for navigating through dynamic indoor environments. Empirical results indicate that robots trained with our ICL-enhanced DRL framework outperform those trained without curriculum learning, highlighting the benefits of structured learning progressions in robotic training.

\end{abstract}

\section{INTRODUCTION}

Human-robot interaction (HRI) has emerged as a pivotal field in robotics research, aiming to bridge the communication gap between humans and autonomous systems. With advancements in natural language processing and machine learning \cite{devlin2019bert} \cite{radford2019language}, enabling mobile robots to comprehend human text instructions has become a promising path for enhancing their flexibility and usability in various domains. In this paper, we delve into the domain of task-based human instructions and their integration with curriculum learning strategies \cite{bengio2009curriculum} \cite{narvekar2020curriculum} to empower mobile robots with enhanced navigational capabilities in complex indoor environments.

Navigating through real-world environments based on human-provided task instructions presents a complex challenge for mobile robots. Unlike traditional navigation tasks where robots follow predefined paths or reach specified waypoints \cite{cai2009information} \cite{raja2015new}, interpreting and executing task-based instructions such as "find the bread and then slice it" requires a deeper understanding of both linguistic nuances and environmental dynamics. This poses a significant obstacle for general deep reinforcement learning (DRL) algorithms \cite{mnih2015human} \cite{schulman2017proximal}, which even often struggle to understand the objectives across diverse tasks and environments.
 
To address these challenges, we propose the utilization of incremental curriculum learning as a key technique for training mobile robots to navigate based on task-oriented human instructions. Curriculum learning offers a structured approach to gradually expose the robot to increasingly complex tasks, enabling it to acquire and refine the necessary skills and knowledge iteratively. By leveraging curriculum learning, we aim to not only improve the task accomplishment rates but also enhance the generalization capabilities of mobile robots across varied tasks and environments.

In this work, we also emphasize the importance of evaluating mobile robot performance in terms of both task accomplishment and generalization metrics. Task accomplishment metrics measure task execution proficiency, while generalization metrics assess its adaptability to adapt and perform effectively in diverse scenarios. Additionally, we conduct a sensitivity analysis of the proposed curriculum design to further assess its robustness across varying training settings. By incorporating these evaluation metrics and sensitivity analysis, we aim to provide comprehensive insights into the effectiveness of our proposed technique.


\begin{figure}[tbp]
    \centering
    \begin{subfigure}[b]{0.48\linewidth}
        \centering
        \includegraphics[width=\linewidth]{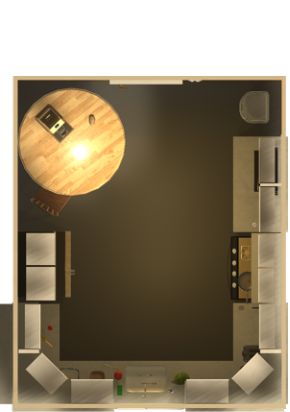}
        \caption{Kitchen Floor 19}
    \end{subfigure}
    \hfill
    \begin{subfigure}[b]{0.48\linewidth}
        \centering
        \includegraphics[width=\linewidth]{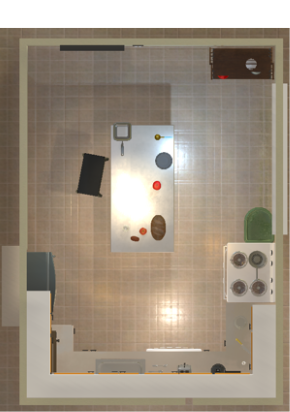}
        \caption{Kitchen Floor 20}
    \end{subfigure}
    \caption{Two examples of AI2-THOR environment simulator available for learning navigation based on human instructions. Our robot model aims to efficiently navigate and accomplish tasks with minimal steps.}
    \label{fig:floorkitchen}
    \vspace{-5mm}
\end{figure}

\section{RELATED WORK}

The development and deployment of mobile robots capable of interpreting human instructions and autonomously navigating through environments represent a critical frontier in artificial intelligence and robotics. The primary focus of this initiative is the integration of DRL with computer vision and natural language processing, allowing robots to understand and act in the physical world based on verbal or written instructions. This section reviews essential works and additional references that contribute to advancements in vision-based navigation, the interpretation of natural language directions, and task-based learning in mobile robotics.

A foundational aspect of mobile robot navigation is their ability to navigate complex environments using visual input. The works by \cite{zhu2016target} and \cite{kulhanek2019vision} have been pivotal in this domain. They propose a DRL-based model enabling robots to navigate towards a visually indicated target in an indoor setting, emphasizing learning scene-specific features. \cite{kulhanek2020visual} further this exploration by demonstrating a system navigating complex 3D mazes, emphasizing the role of auxiliary tasks in enhancing navigation. These studies underscore the critical function of deep learning in enabling robots to interpret visual data for navigation, setting a foundation for further research into task-based instruction.

Beyond visual navigation, enabling robots to follow natural language instructions is crucial for effective human-robot interaction. \cite{shah2018follownet} present "FollowNet", a model that interprets and follows complex natural language directions in real-time, combining the strengths of DRL with natural language processing. This approach enhances the robot's navigational capabilities and its practical utility in scenarios where verbal instructions are commonly encountered.

The concept of curriculum learning is central to teaching robots to perform a wide range of tasks based on human instructions. The ALFRED benchmark by \cite{shridhar2020alfred} offers a comprehensive framework for evaluating models that interpret sequential instructions for completing household tasks. This benchmark challenges robots to execute actions based on visual observations and textual descriptions, pushing the boundaries for more sophisticated models capable of understanding and acting upon complex instructions.

Expanding on these foundational works, subsequent research has delved into memory-augmented networks for improved path planning by \cite{parisotto2017neural} and the use of multimodal data for understanding instructions in noisy environments by \cite{thomason2015learning}. Moreover, the application of curriculum learning \cite{gunukula2023miracle}, where robots are gradually exposed to more complex tasks, mirrors educational strategies in humans, suggesting a cross-disciplinary approach to robot training.

\section{AI2THOR FRAMEWORK}

\begin{figure}[tbp]
    \centering
    \includegraphics[width=\linewidth]{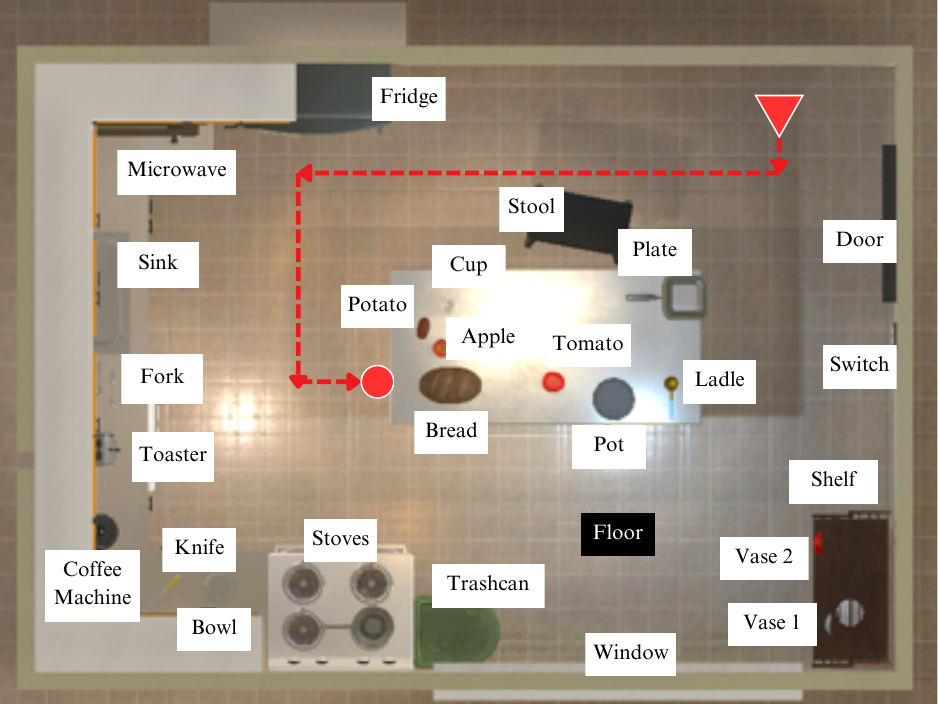}
    \caption{Bird's-eye view kitchen room with variety of objects in AI2-THOR environment. The floor can also be identified during the navigation.}
    \label{fig:floorbird}
    \vspace{-5mm}
\end{figure}

To address the challenge of training and evaluating mobile robots to follow task-based human instructions within a 3D environment, we employed the AI2-THOR framework \cite{kolve2017ai2thor}. This platform merges the UNITY 3D engine's detailed physics simulation which offers a realistic environment (refer to figure \ref{fig:floorkitchen}) where robots can execute complex behaviors. AI2-THOR’s range of indoor environments, from kitchens to living rooms, furnished with interactable objects, provides a diverse testing ground. Such an environment is crucial for robots to learn tasks as intricate as navigating to objects, as depicted in figure \ref{fig:floorbird}, and manipulating them, reflecting real-world complexity.

The utility of AI2-THOR extends beyond mere simulation, facilitating curriculum learning where robots progressively tackle tasks of increasing complexity. This approach mirrors human learning, ensuring robots develop skills systematically, from basic navigation and object interaction to executing complicated, multi-step tasks based on verbal instructions. The framework's comprehensive environments enable structured learning, making it an invaluable tool for refining the adaptability and autonomy of mobile robots in performing a wide range of household tasks.

Our research, inspired by and contributing to a body of work leveraging AI2-THOR, including visual question answering by \cite{gordon2018iqa}, robotic grasping techniques by \cite{batra2020rearrangement}, and smart episodic by \cite{pashevich2021episodic}, showcases the framework's broad applicability. These precedents highlight AI2-THOR's role in advancing AI and robotics research, proving its effectiveness in training models for complex interaction and understanding within a physically plausible virtual world. Through our focused application of AI2-THOR, we aim to improve the capability of mobile robots to comprehend and execute complex task-based instructions, thereby pushing the boundaries of robot interaction with humans.

\section{TASK-BASED ROBOT MODEL}

In this section, we first define our formulation for task-based visual navigation. Then we describe our deep neural network designed as the robot model (agent) for this work.

\subsection{Problem Statement}

\begin{figure}[tbp]
    \centering
    \includegraphics[width=0.8\linewidth]{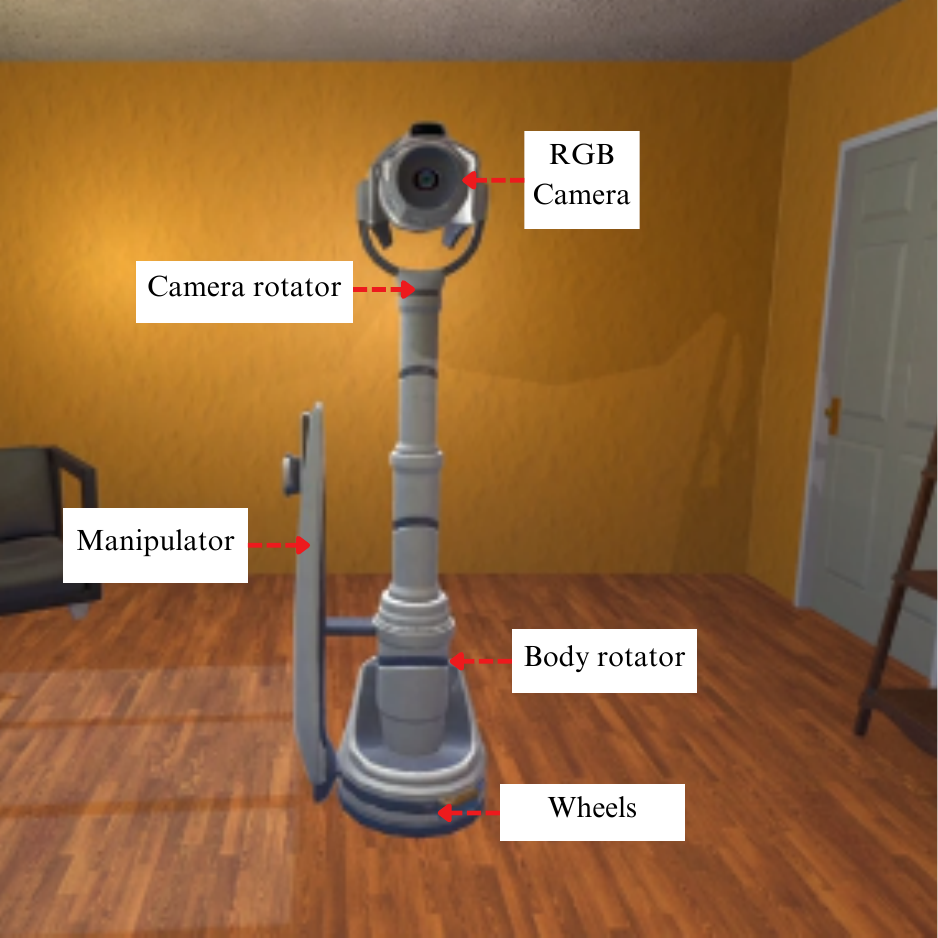}
    \caption{Snapshot of the mobile robot simulated within the AI2-THOR framework.}
    \label{fig:robotlook}
    \vspace{-5mm}
\end{figure}

Our goal is to find the most effective sequence of actions necessary to fulfill specific tasks outlined by human instructions, such as "find the bread, and then slice it". To achieve this, we developed a deep reinforcement learning model capable of processing both an RGB image representing the current observation and a textual description of the task objective. The robot model generates a 3D action output, such as moving forward or rotating left, based on the inputs provided. Figure \ref{fig:robotlook}  illustrates the detailed robot model and its features. 

\subsection{Problem Formulation}

Vision-based and text-based robot navigation requires the translation of sensory and textual signals into actionable motion commands. However, traditional deep reinforcement learning (DRL) approaches often overlook the complexity of integrating high-dimensional visual inputs and textual instructions simultaneously. Recent advancements in DRL have introduced end-to-end learning frameworks capable of seamlessly processing both pixel information and textual data. However, integrating information from diverse modalities like vision and text into a unified representation remains challenging, requiring effective capture of complementary information from each modality to facilitate learning. In this context, the policy must learn to extract relevant details from both vision and text and make informed decisions based on this combined knowledge.

\begin{figure}[h!]
    \centering
    \begin{subfigure}[b]{0.30\linewidth}
        \centering
        \includegraphics[width=\linewidth]{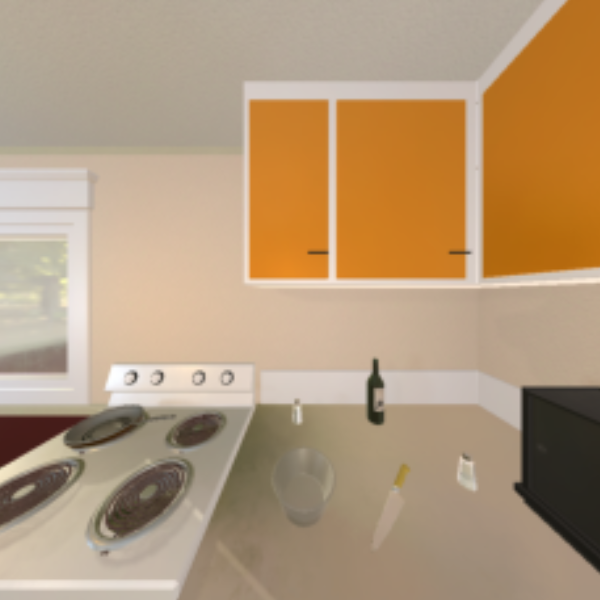}
        \caption{A bowl on the countertop}
    \end{subfigure}
    \hfill
    \begin{subfigure}[b]{0.30\linewidth}
        \centering
        \includegraphics[width=\linewidth]{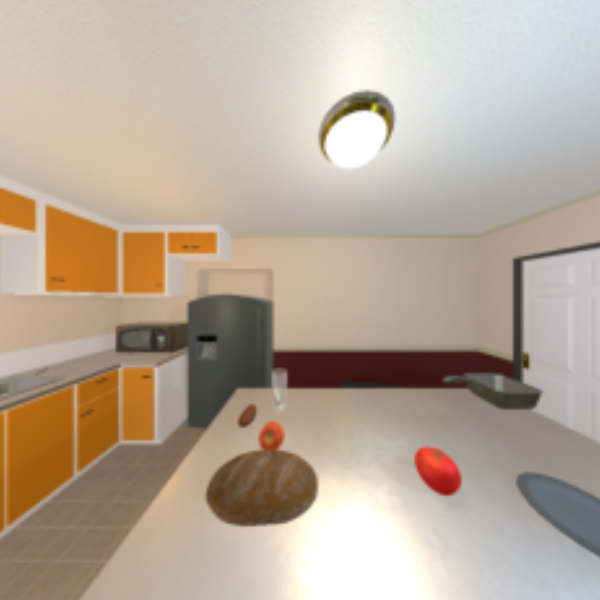}
        \caption{A bread on the table}
    \end{subfigure}
    \hfill
    \begin{subfigure}[b]{0.30\linewidth}
        \centering
        \includegraphics[width=\linewidth]{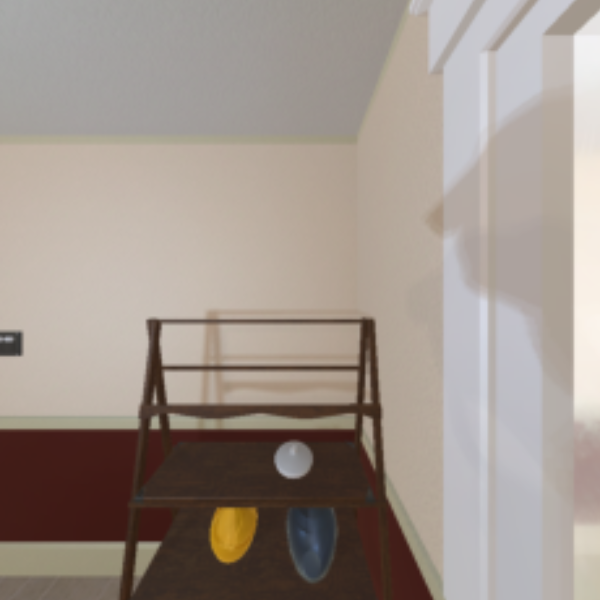}
        \caption{A vase on the shelf}
    \end{subfigure}
    \caption{Our robot model observes RGB visual image data. Within a single observation, multiple objects in the kitchen room can be identified by the agent.}
    \label{fig:objkitchen}
    \vspace{-2mm}
\end{figure}

In the context of mobile robot navigation, especially in environments where human interaction is required, such as home settings, the challenge of integrating high-dimensional sensory inputs, such as RGB images (as illustrated in figure \ref{fig:objkitchen}), and textual inputs becomes increasingly apparent. In such scenarios, robots must comprehend and execute complex instructions provided in natural language, such as "find the bread, take it, go to the fridge, and place the bread inside". This task involves not only understanding the semantic meaning of the instructions but also extracting actionable commands from them. However, it is technically feasible to train a model to learn a stochastic policy function that takes two inputs: a representation of the current state and a representation of human instructions in text form. Subsequently, this model generates a probability distribution across the action space.

\subsection{Curriculum Design}

Curriculum design in the context of training the agent in the AI2-THOR simulator with multimodal inputs involves carefully structuring the learning process to facilitate effective skill acquisition and adaptation over time. By gradually increasing the complexity of tasks and providing appropriate rewards and guidance, we consider that curriculum learning can help the agent learn more efficiently and robustly in complex environments.

\begin{table}[tbp]
\centering
\caption{Example of Possible Human Instructions at Kitchen Room (AI2-THOR Simulator)}
\begin{tabular}{c | p{6cm}} 
 \hline
 No. & Human Instruction \\ [1.5ex]
 \hline\hline
 1 & "Find the bread, take it, go to the fridge, and place the bread inside." \\ [1.5ex]
 2 & "Search for the apple, grab it, move to the table, and put the apple on the plate." \\ [1.5ex]
 3 & "Identify the stove, turn it on, go to the pot, and pour water into it." \\ [1.5ex]
 4 & "Search for the cup, take it, go to the sink, and fill the cup with water." \\ [1.5ex]
 5 & "Spot the refrigerator, open it, find the milk carton, and take it out." \\ [1.5ex]
 6 & "Locate the cereal box, pick it up, then find the bowl and pour cereal into it." \\ [1.5ex] 
 7 & "Identify the fruit bowl, pick an apple, wash it in the sink, and place it on the cutting board." \\ [1.5ex] 
 8 & "Spot the coffee machine, turn it on, find a mug, and place it under the coffee dispenser." \\ [1.5ex] 
 \hline

\end{tabular}
\label{tab:huminstructions}
\vspace{-2mm}
\end{table}

In this work, we address the challenge posed by the diverse range of objects and possibly executed actions present within a room, which complicates task execution for agents. To tackle this challenge, we propose an incremental curriculum learning (ICL) approach. Human instructions in environments like kitchens often involve multiple object recognitions and corresponding actions, as depicted in Table \ref{tab:huminstructions}. Our curriculum is designed to systematically introduce and master these tasks by breaking them down into manageable steps. Table \ref{tab:exainstructions} provides a clear example of how the complex instruction "find the bread, take it, go to the fridge, and place the bread inside" can be deconstructed into four stages, forming the basis of our curriculum design.

Before passing into the robot model, the complete human instruction is first divided into several stages. Initially, the instruction is broken down into several stages, each representing a distinct action or task. Subsequently, stop words are utilized to filter out unnecessary words, enhancing the relevance of the remaining content. Finally, each word in the processed instruction is mapped to its corresponding feature representation using GLOVE \cite{pennington2014glove} word embeddings, facilitating the generation of a comprehensive input representation for the model. This multi-step preprocessing ensures that the instruction is effectively transformed into a structured and informative input for the robot model. It is important to note that in this work, we simplify the preprocessing step by excluding popular attention mechanisms \cite{vaswani2017attention}, as they are beyond the scope of this research. 

\begin{table}[tbp]
\centering
\caption{Example of Instructions at Different Stages}
\begin{tabular}{c | p{6cm} } 
 \hline
 Stage & Curriculum-based Instruction \\ [1.5ex] 
 \hline\hline
 1 & "Find the bread" \\  [1.5ex]
 2 & "Find the bread, take it" \\ [1.5ex]
 3 & "Find the bread, take it, go to the fridge" \\ [1.5ex]
 4 & "Find the bread, take it, go to the fridge, and place the bread inside" \\ [1.5ex] 
 \hline
\end{tabular}
\label{tab:exainstructions}
\vspace{-2mm}
\end{table}

\subsection{Learning Setup}

Before introducing our model, we first describe the key concepts of the deep reinforcement learning setup implemented in this work including action space, observations and goals, and reward design.

\begin{itemize}

\item Action Space: Our model is trained using command-level actions, covering five tasks: moving forward, rotating left, rotating right, picking up an object, and throwing an object. We use a step length of 0.25 meters and a turning angle of 90 degrees, effectively discretizing the scene space into a grid-world representation.

\item Observations and goals: The observation comprises images taken by the agent's RGB camera from a first-person perspective, whereas the goal originates from text-based human instructions. The benefit of employing textual data for goal description lies in its adaptability for task specification and its enhancement of interaction between humans and robots.

\item Reward design: Our primary goal is to minimize the trajectory length required to accomplish tasks specified by human instructions. As such, we provide a goal-reaching reward (ranging from 5.0 to 20.0 based on the curriculum stage) upon task completion. Additionally, to encourage shorter trajectories, we apply a small penalty (-0.05) as an immediate reward.

\end{itemize}

\subsection{Model}

Our main focus is on training a task-driven policy function using deep reinforcement learning. We propose a deep neural network architecture as a non-linear function approximator to determine the action \( a \) at time \( t \). Let \( \pi_{\theta} \) denote the deep neural network (DNN), where \( \theta \) represents the network's parameters. We can express the action \( a \) at time \( t \) as a function of the current state \( s_t \) and the network parameters \( \theta \). Mathematically, this can be represented as:

\begin{equation} 
a_t = \pi_{\theta}(s_t)
\label{eq:policy_function}
\end{equation}

where policy function \( \pi_{\theta} \) maps the current state \( s_t \) to an action \( a_t \) at time \( t \) using the parameters \( \theta \) of the neural network. However, given the numerous real-world goals resulting from the wide variety of objects, actions, and locations, it is preferable to learn a projection that can transform these inputs into a shared embedding space, enabling the model to represent information from both modalities that facilitate the integration of multiple modalities and assists in decision-making processes.


In the scenario where the agent receives both visual input and text input, the task becomes inherently multimodal. The agent must effectively integrate information from both modalities to make informed navigational decisions within the environment. The visual input provides the agent with information about its surroundings, including the layout of the room, the locations of objects, and the overall scene geometry. On the other hand, the text input, such as human instructions like "find bread and then take it", provides the agent with high-level goals or objectives. 

To address the multimodal challenge, we developed a Multimodal Deep Q Network (MDQN). The neural network architecture is designed to process two types of input data: images and text. The image input is subjected to feature extraction through a Convolutional Neural Network (CNN), resulting in the embedding \(E_{\text{image}}\). Similarly, the text input is handled by a Long Short-Term Memory (LSTM) network, yielding the embedding \(E_{\text{text}}\).

\begin{equation}
\begin{aligned}
& \quad E_{\text{image}} = \text{CNN}(\text{image}) \\
& \quad E_{\text{text}} = \text{LSTM}(\text{text}) \\
& \quad E_{\text{concat}} = \text{Concatenate}(E_{\text{image}}, E_{\text{text}}) \\
& \quad H_1 = \text{ReLU}(W_1 E_{\text{concat}} + b_1) \\
& \quad H_2 = \text{ReLU}(W_2 H_1 + b_2) \\
& \quad \text{Q-values} = W_3 H_2 + b_3 \\
\end{aligned}
\end{equation}

These embeddings are then concatenated into a unified representation, \(E_{\text{concat}}\), allowing the network to incorporate both image and text information, similar to the approach used in previous works \cite{donahue2015long} \cite{malinowski2016ask}. The concatenated representation is subsequently passed through two fully connected layers, each followed by a Rectified Linear Unit (ReLU) activation function. In these layers, weights (\(W_1\), \(W_2\), \(W_3\)) control input transformations, while biases (\(b_1\), \(b_2\), \(b_3\)) offer flexibility in output adjustment. The hidden representations \(H_1\) and \(H_2\) are produced after applying the ReLU activation function. Finally, the output layer computes the Q-values by DQN algorithm \cite{mnih2013playing}, representing the expected future rewards for each possible action, facilitating decision-making process in reinforcement learning scenarios.

\subsection{Network Architecture}

We present the network for processing both visual and textual inputs in task-oriented learning scenarios. Our model incorporates a pretrained ResNet18 \cite{he2016deep} for handling visual information, with the ResNet parameters held fixed during training. The textual data is processed through an LSTM model equipped with GloVe embeddings, utilizing a hidden dimension of 64. Meanwhile, the multimodal architecture concatenates the extracted features from both models, producing a projected embedding vector. This vector passes through three fully connected layers with dimensions of 640, 512, and 256, respectively. The output comprises 3-5 policy outputs (with respect to the curriculum stage), and a single value output. We trained this network using a shared RMSProp optimizer with a learning rate of \(1 \times 10^{-4}\).

\section{EXPERIMENTS}

In this section, we evaluate our trained navigating agent's effectiveness in executing task-based human instructions and assessing its achievability and generalization capability. We compare our model with baseline models trained without curriculum learning, highlighting its efficient task accomplishment. Furthermore, we explore different reward designs to mitigate catastrophic forgetting and extend our model's capability to handle a wider range of objects. Sensitivity analysis regarding curriculum learning strategy performance is also presented.

\subsection{Comparison with Baseline}

\begin{figure}[h!]
    \centering
    \begin{subfigure}[b]{0.48\linewidth}
    \includegraphics[width=\linewidth]{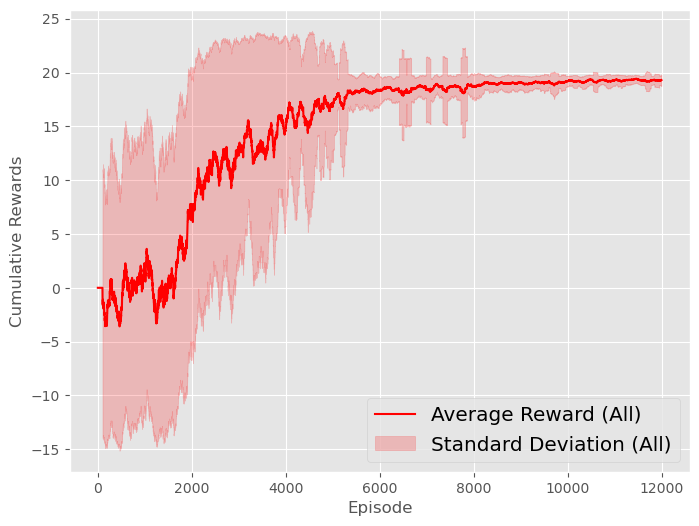}
    \caption{Stage 1}
    \end{subfigure}
    \begin{subfigure}[b]{0.48\linewidth}
    \includegraphics[width=\linewidth]{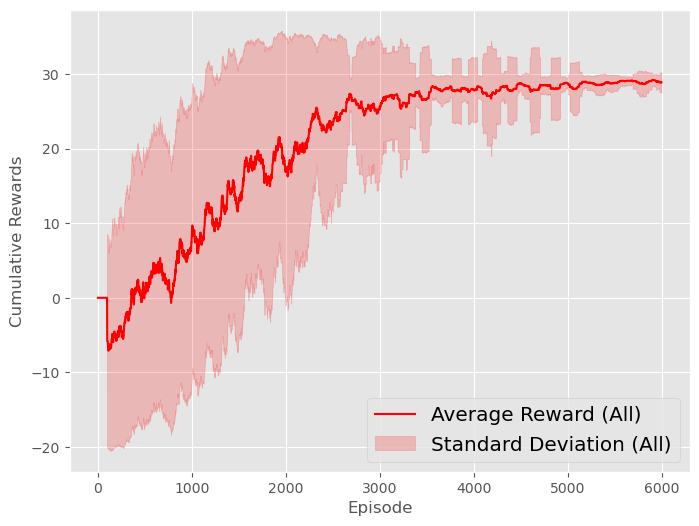}
    \caption{Stage 2}
    \end{subfigure}
    
    \begin{subfigure}[b]{0.48\linewidth}
    \includegraphics[width=\linewidth]{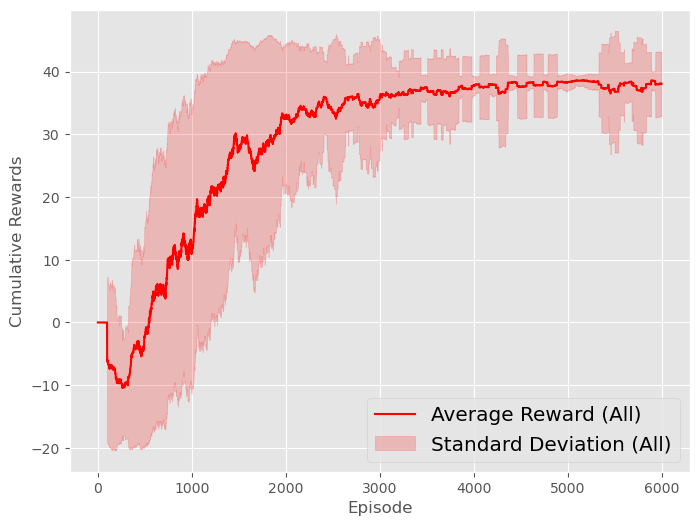}
    \caption{Stage 3}
    \end{subfigure}
    \begin{subfigure}[b]{0.48\linewidth}
    \includegraphics[width=\linewidth]{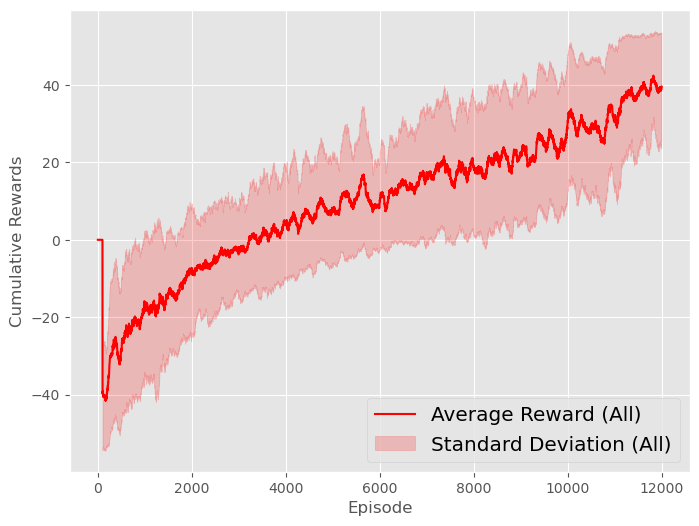}
    \caption{Stage 4}
    \end{subfigure}
    
    \caption{Training results from each stage indicate that our model successfully learns to execute the instructions after completing four stages of distinct training processes.}
    \label{fig:allstagesub}
    \vspace{-2mm}
\end{figure}

We implemented our models using PyTorch Deep Learning Framework \cite{paszke2019pytorch} and train them on an NVIDIA GeForce RTX 3050. In our experiment, we utilized incremental curriculum learning (ICL) to train an agent to navigate and interact in a kitchen room environment, focusing solely on 3 objects: a bowl, bread, and vase. The training initiated from stage 1, where the agent was tasked with the simplest objective of heading to the objects in the room and progressed through 12,000 episodes. At the start of each episode, the robot's state was randomly initialized. As depicted in figure \ref{fig:allstagesub}, the curriculum learning approach enabled the agent to effectively learn navigation and locate objects.

We implemented linear decay for epsilon, an exploration-exploitation trade-off parameter, which ensured that the exploration rate gradually decreased as the episode increased. This approach encouraged the agent to increasingly exploit learned knowledge as training progressed. Furthermore, examining the learning curves of stages 2 and 3 showed a similar trend, indicating that the agent effectively utilized transfer learning from previous stages. However, stage 4 presented a challenge due to the complexity of handling complete instructions, often comprising lengthy sentences. Despite this difficulty, the agent demonstrated the capability to learn and execute meaningful instructions, such as "find the bread, take it, go to the fridge, and place the bread inside". This observation underscores the agent's ability to acquire knowledge across varying levels of task complexity.

\begin{figure}[h!]
    \centering
    \includegraphics[width=\linewidth]{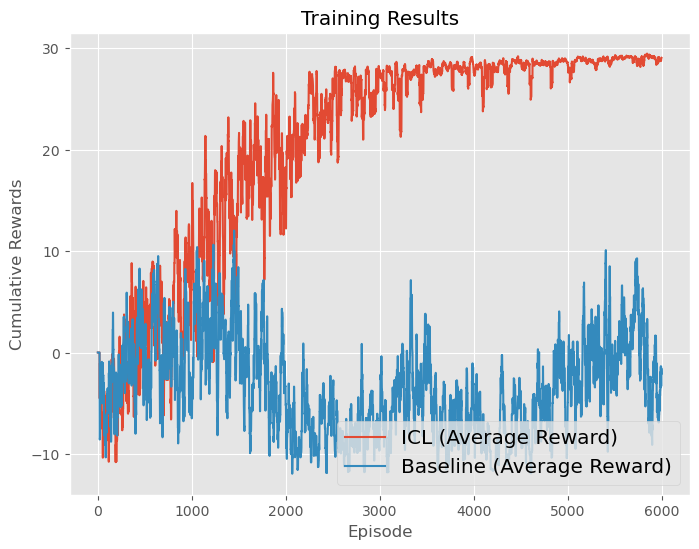}
    \caption{Comparison of training results between agents trained with instructions from stage 2 using Incremental Curriculum Learning (ICL) and their baseline counterparts (trained without ICL).}
    \label{fig:iclbaseline}
    \vspace{-2mm}
\end{figure}

As illustrated in figure \ref{fig:iclbaseline}, the absence of curriculum learning makes maintaining the achievability of the task challenging, as the agent lacks a structured approach to learning. In contrast, agents trained with ICL demonstrate enhanced learning capabilities, as they are systematically exposed to progressively more challenging tasks. Without curriculum learning, the agent struggles to learn how to accomplish the task described in human instructions, highlighting the importance of providing a structured learning framework to facilitate effective task accomplishment.

\subsection{Positive vs Neutral Reward}

\begin{figure}[h!]
    \centering
    \includegraphics[width=\linewidth]{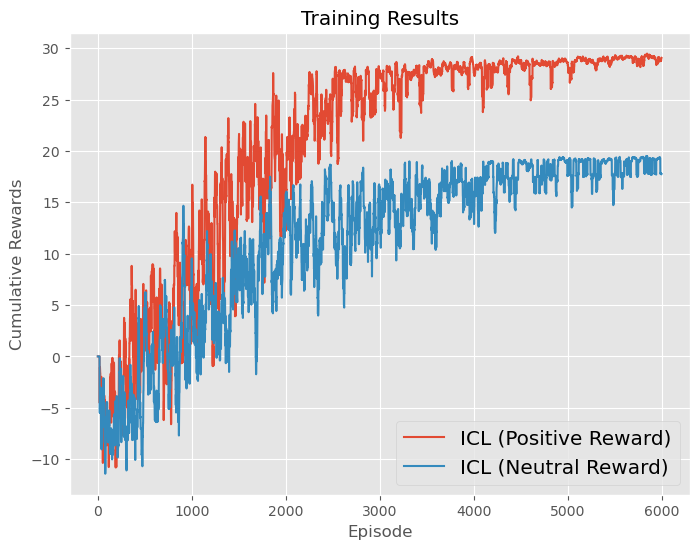}
    \caption{Comparison of training results between agents trained with instructions from stage 2 using positive reward and their baseline counterparts (neutral reward). Both use curriculum learning.}
    \label{fig:neutralreward}
    \vspace{-5mm}
\end{figure}

Introducing positive reward for tasks that have been mastered from previous stages in curriculum learning can be beneficial under certain circumstances. By consistently rewarding the successful completion of mastered tasks, positive reward shaping helps solidify these behaviors. This strategy is particularly effective in mitigating catastrophic forgetting by encouraging the retention of existing knowledge and skills \cite{anca2023achieving} \cite{kirkpatrick2017overcoming}. However, as depicted in the figure \ref{fig:neutralreward}, the inclusion of positive reward does not significantly affect the learning process. This is likely because the task design outlined in the curriculum supports the mastery of tasks and prevents quick transitions to more complex tasks. As shown in the curve, the utilization of positive reward only slightly enhances learning convergence compared to neutral reward.

\subsection{Generalization Capability}

\begin{figure}[h!]
    \centering
    \includegraphics[width=\linewidth]{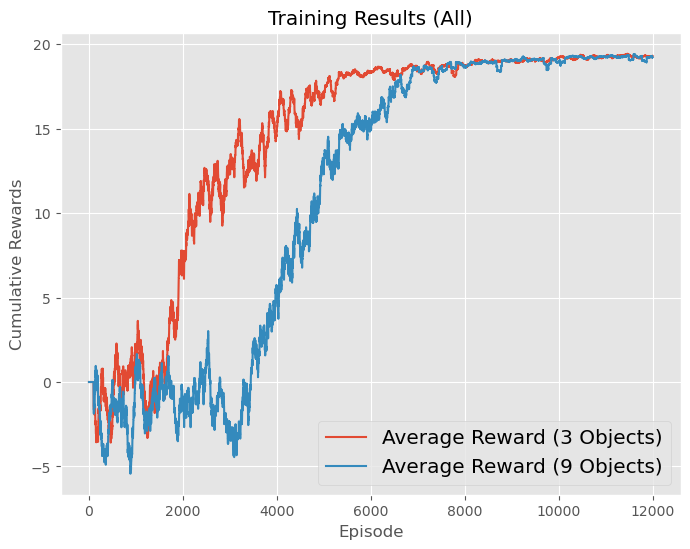}
    \caption{Comparison of training results between agents trained to navigate to 3 objects and those trained to navigate to 9 objects in the same kitchen room.}
    \label{fig:icl3-9objs}
    \vspace{-2mm}
\end{figure}

The learning curves depicted in figure \ref{fig:icl3-9objs} demonstrates the generalization capability of our agent across varying levels of task complexity. Initially trained to handle 3 objects in the kitchen room, our model exhibits adaptability when tasked with learning to manage 9 objects using the same model architecture. The training process restarts, with the model's weights and biases randomly initialized. Despite the increased complexity of handling a broader range of objects, the learning curves indicate that our agent maintains its ability to accomplish tasks effectively. As the number of objects increases, the agent's need for exploration becomes more evident, particularly given the random initiation of its respawn location. This robust learning mechanisms to navigate and interact within the environment, showcasing the model's capacity to generalize across diverse scenarios.

\begin{figure}[h!]
    \centering
    \includegraphics[width=\linewidth]{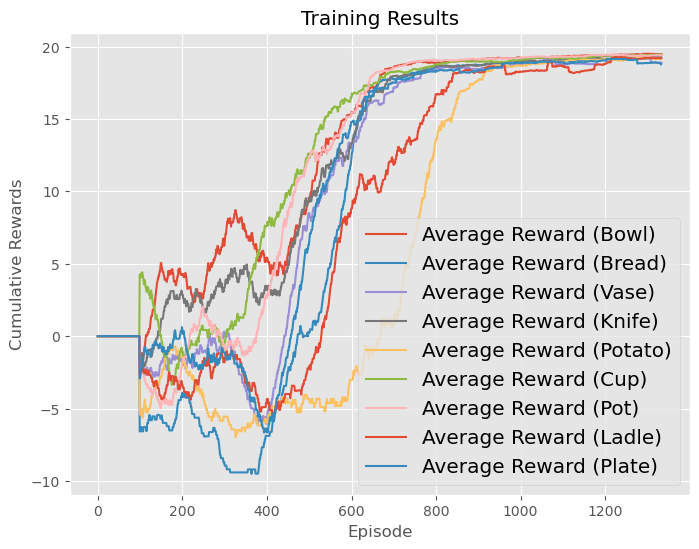}
    \caption{Each curve represents the agent's success in navigating to and locating each item. There are indications that the agent needs further exploration to successfully identify more objects.}
    \label{fig:icl9objs}
    \vspace{-5mm}
\end{figure}

The decision to incorporate the 9 objects to evaluate the model's generalization is based on the various locations within the kitchen environment. Figure \ref{fig:icl9objs} illustrates the agent successfully navigating and locating these objects. However, few objects may require additional exploration to be correctly identified, likely due to objects such as potatoes, ladles, and plates clustered together on the living table presenting a challenge for the agent to distinguish between them. The agent's ability to successfully navigate and locate many objects across diverse locations demonstrates its promising performance in real-world scenarios.

\subsection{Sensitivity Analysis}

\begin{figure*}[tbp]
  \centering
  \begin{subfigure}[b]{0.32\linewidth}
    \includegraphics[width=\linewidth]{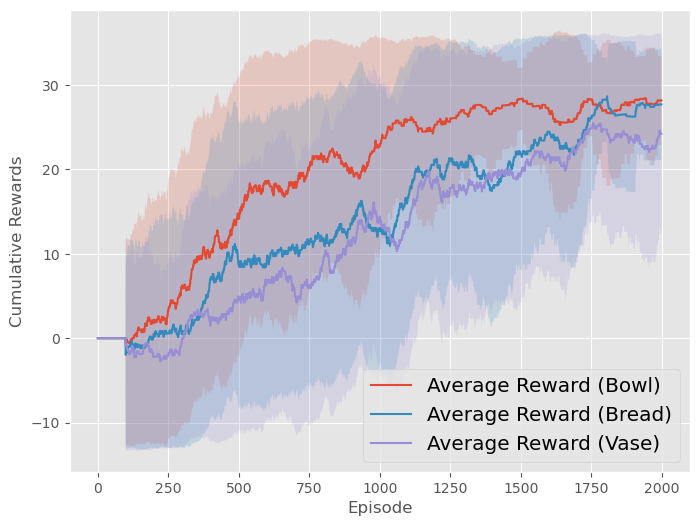}
    \caption{Maxtime 100}
  \end{subfigure}
  \begin{subfigure}[b]{0.32\linewidth}
    \includegraphics[width=\linewidth]{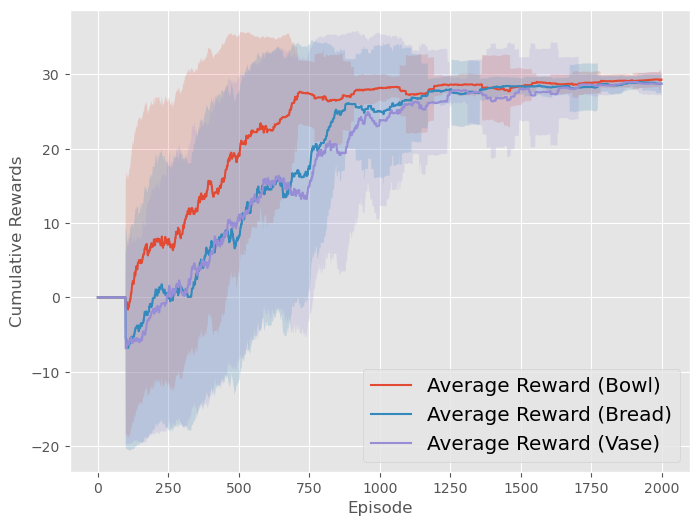}
    \caption{Maxtime 200}
  \end{subfigure}
  \begin{subfigure}[b]{0.32\linewidth}
    \includegraphics[width=\linewidth]{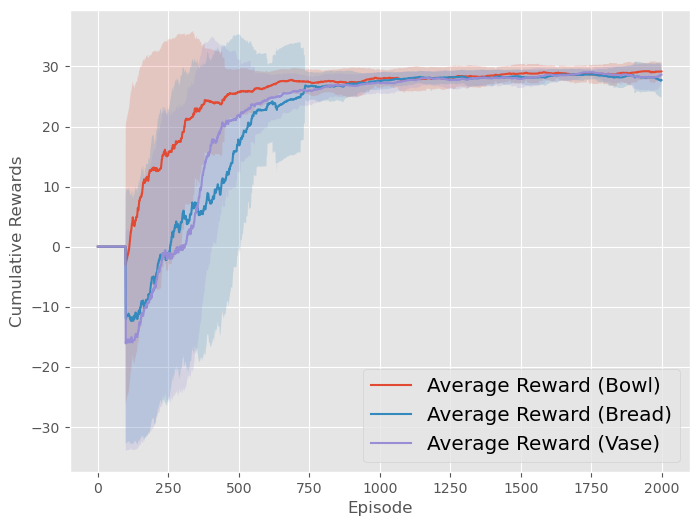}
    \caption{Maxtime 400}
  \end{subfigure}
  
  \begin{subfigure}[b]{0.32\linewidth}
    \includegraphics[width=\linewidth]{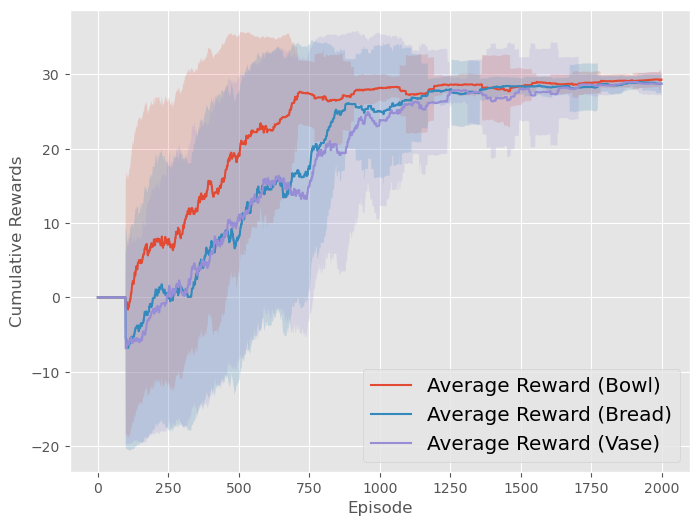}
    \caption{Epsilon 0.90}
  \end{subfigure}
  \begin{subfigure}[b]{0.32\linewidth}
    \includegraphics[width=\linewidth]{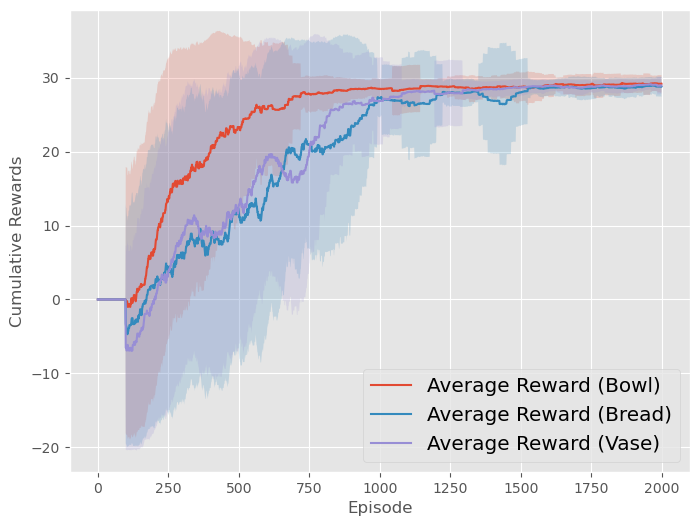}
    \caption{Epsilon 0.75}
  \end{subfigure}
  \begin{subfigure}[b]{0.32\linewidth}
    \includegraphics[width=\linewidth]{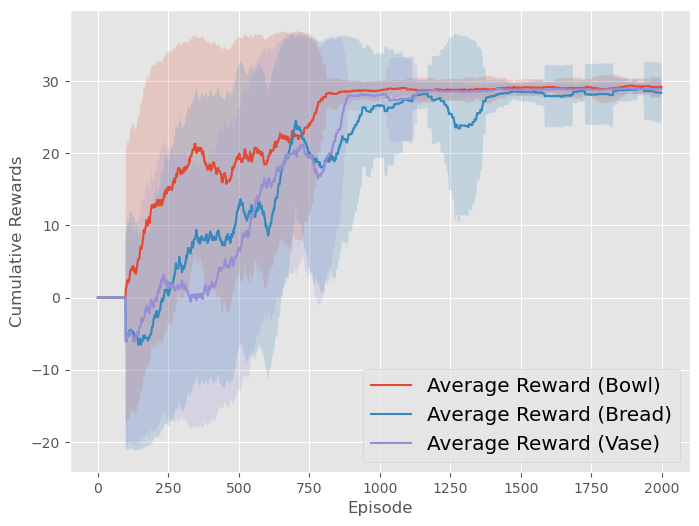}
    \caption{Epsilon 0.50}
  \end{subfigure}

  \begin{subfigure}[b]{0.32\linewidth}
    \includegraphics[width=\linewidth]{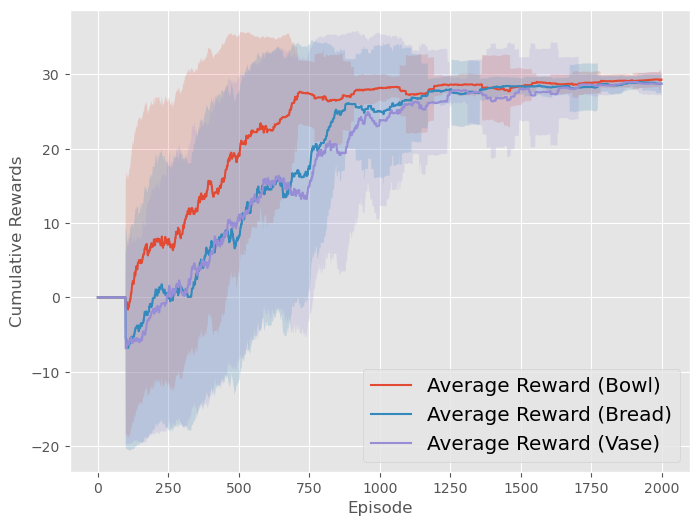}
    \caption{Reward Incremental}
  \end{subfigure}
  \begin{subfigure}[b]{0.32\linewidth}
    \includegraphics[width=\linewidth]{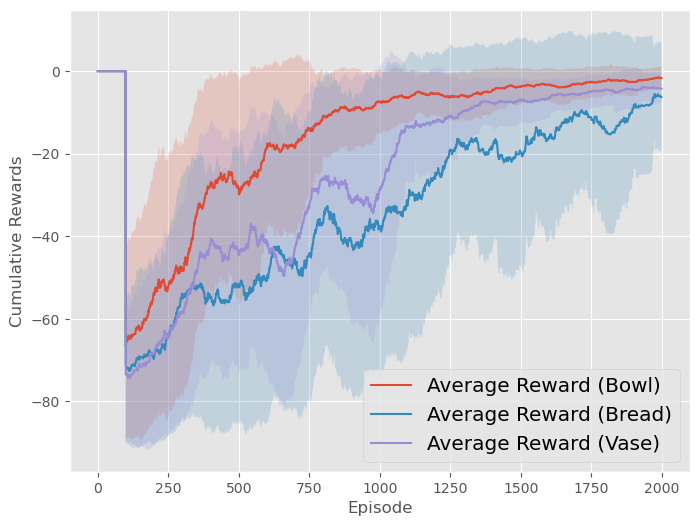}
    \caption{Reward Norm}
  \end{subfigure}
  \begin{subfigure}[b]{0.32\linewidth}
    \includegraphics[width=\linewidth]{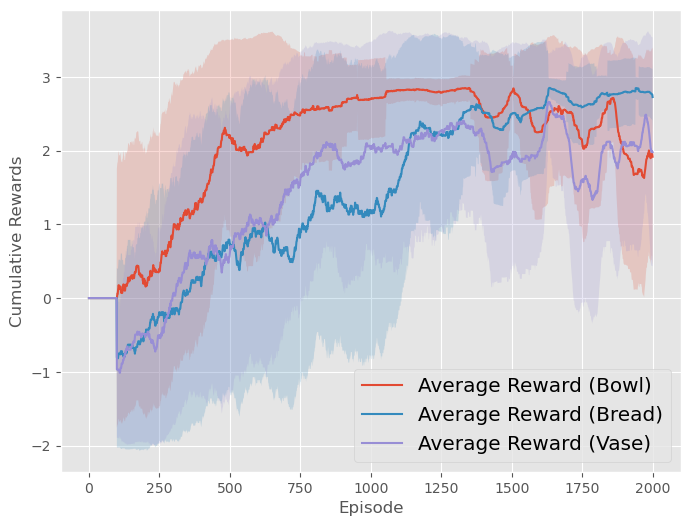}
    \caption{Reward Div10}
  \end{subfigure}
  
  \caption{Results of sensitivity analysis for hyperparameters including maxtime, epsilon, and reward scaling during training. To conduct a sensitivity analysis, we focused on the learning process of the agent at stage 2 of the curriculum.}
  \label{fig:stage2sensitivity}
  \vspace{-2mm}
\end{figure*}

To deepen our understanding of the model's robustness and reliability concerning variations in its hyperparameters related to curriculum learning, we conducted a sensitivity analysis. We examined important hyperparameters to consider when implementing curriculum learning, such as max time, epsilon, and reward scaling. Max time represents the step limit the agent can perform during an episode, regardless of whether it completes the task. Epsilon represents the rate of exploration, and reward scaling is the scalar value scale that contributes to reinforcing the agent's learning. After conducting the sensitivity analysis, we gained insights into how variations in these hyperparameters affect the model's performance and learning dynamics. 

As illustrated in figure \ref{fig:stage2sensitivity}, increasing the agent's step limit (maxtime) resulted in better performance, indicating the importance of allowing more time for task completion. Reducing epsilon resulted in improved results in term of learning speed, suggesting that less exploration enhances learning by depending more on existing knowledge. Additionally, utilizing reward incremental, which adds positive rewards for task completion, significantly outperformed other reward scaling methods (smaller number of rewards by normalization or dividing by 10). This provides us with a good indication that in this work, we can design the reward intuitively without worrying much about its scale.

We also observed that regardless of the object's location, the model trained using curriculum learning effectively leveraged its existing knowledge to successfully accomplish tasks. However, it became apparent that the selection of hyperparameters played a crucial role in the model's performance. These findings emphasize the significance of hyperparameters tuning such as maxtime, epsilon, and reward scaling in optimizing agent performance and learning efficiency.

\section{CONCLUSIONS}

We introduced a deep reinforcement learning (DRL) framework for visual navigation guided by task-based human instructions. Current DRL methods typically rely solely on visual images, which may not always be optimal for enhancing human-machine interaction. This work represents a step towards addressing this gap by prioritizing the use of the intuitive curriculum learning technique. We addressed this gap by focusing on the integration of incremental curriculum learning (ICL). Specifically, we tackled challenges associated with learnin g from both visual and textual inputs, especially when the desired goal is non-trivial such as human instruction. Additionally, we conducted a sensitivity analysis to deepen our understanding of the effectiveness of this technique in improving agent learning and its implementation.

Our experiments demonstrated the effectiveness of our method with curriculum learning strategy in terms of task accomplishment and generalization capability, while the approach without it failed to succeed. We also observed that our curriculum along with the incremental reward design can help assist in mitigating the risk of catastrophic forgetting when implementing the strategy. Additionally, we examined the potential extension of our model to accommodate a wider variety of objects and destinations during navigation while accomplishing requested tasks. 

Our future work involves enhancing the capacity and flexibility of processing text-based instructions by implementing attention mechanisms, which can associate important words with the visual observations acquired by the agent. Additionally, other curriculum learning methodologies will also be explored and we aim to develop a more advanced model capable of recognizing unseen instructions and effectively navigating through similar or even complex environments.

\section*{ACKNOWLEDGMENT}

We would like to express our gratitude to the Ministry of Education, Culture, Sports, Science and Technology (MEXT) of Japan for their support of this research through the provision of a scholarship under the special program known as Human-Centered AI at the University. We also would like to thank Victor Kich for his helpful comments.

\end{document}